\documentclass[conference]{IEEEtran}
\IEEEoverridecommandlockouts

\usepackage{cite}
\usepackage{amsmath,amssymb,amsfonts}
\usepackage{algorithmic}
\usepackage{graphicx}
\usepackage{textcomp}
\usepackage{xcolor}
\usepackage{url}
\def\BibTeX{{\rm B\kern-.05em{\sc i\kern-.025em b}\kern-.08em
    T\kern-.1667em\lower.7ex\hbox{E}\kern-.125emX}}

\usepackage{multirow}
\usepackage{booktabs}

\usepackage{tikz}
\usepackage{subcaption}
\usetikzlibrary{arrows.meta,positioning,shapes.geometric}
\usetikzlibrary{shapes, arrows, positioning, calc}

\definecolor{msgreen}{RGB}{40,140,80}
\definecolor{lsorange}{RGB}{220,140,20}
\definecolor{green}{RGB}{0, 128, 0}
\definecolor{amber}{RGB}{255, 191, 0}
\definecolor{greensoft}{RGB}{220, 255, 220}
\definecolor{ambersoft}{RGB}{255, 245, 220}

\begin{document}

\title{MRMS-Net and LMRMS-Net: Scalable Multi-Representation Multi-Scale Networks for Time Series Classification}

\author{\IEEEauthorblockN{1\textsuperscript{st} Celal Alagöz}
\IEEEauthorblockA{\textit{Computer Engineering} \\
\textit{Sivas Bilim ve Teknoloji Üniversitesi}\\
Sivas, Türkiye \\
0000-0001-9812-1473} 
\and
\IEEEauthorblockN{2\textsuperscript{nd} Mehmet Kurnaz}
\IEEEauthorblockA{\textit{Computer Engineering} \\
\textit{Sivas Bilim ve Teknoloji Üniversitesi}\\
Sivas, Türkiye \\
0009-0003-1303-1623} 
\and
\IEEEauthorblockN{3\textsuperscript{rd} Farhan Aadil}
\IEEEauthorblockA{\textit{Computer Engineering} \\
\textit{Sivas Bilim ve Teknoloji Üniversitesi}\\
Sivas, Türkiye \\
0000-0001-8737-2154} 
}

\maketitle

\begin{abstract}
Time series classification (TSC) performance depends not only on architectural design but also on the diversity of input representations. In this work, we propose a scalable multi-scale convolutional framework that systematically integrates structured multi-representation inputs for univariate time series. 

We introduce two architectures: MRMS-Net, a hierarchical multi-scale convolutional network optimized for robustness and calibration, and LMRMS-Net, a lightweight variant designed for efficiency-aware deployment. In addition, we adapt LiteMV—originally developed for multivariate inputs—to operate on multi-representation univariate signals, enabling cross-representation interaction.

We evaluate all models across 142 benchmark datasets under a unified experimental protocol. Critical Difference (CD) analysis confirms statistically significant performance differences among the top models. Results show that LiteMV achieves the highest mean accuracy, MRMS-Net provides superior probabilistic calibration (lowest NLL), and LMRMS-Net offers the best efficiency–accuracy tradeoff. Pareto analysis further demonstrates that multi-representation multi-scale modeling yields a flexible design space that can be tuned for accuracy-oriented, calibration-oriented, or resource-constrained settings.

These findings establish scalable multi-representation multi-scale learning as a principled and practical direction for modern TSC. Reference implementation of MRMS-Net and LMRMS-Net is available at: \url{https://github.com/alagoz/mrmsnet-tsc}

\end{abstract}

\section{Introduction}

TSC has witnessed substantial progress with the emergence of deep convolutional and transformer-based architectures. Despite these advances, two fundamental aspects remain underexplored in a unified manner: (i) the role of structured representation diversity, and (ii) the trade-off between accuracy, calibration, and computational efficiency at scale.

Most existing deep TSC models operate on raw time-domain inputs, implicitly expecting the network to learn all relevant transformations internally. However, classical signal processing suggests that complementary representations—such as derivatives, frequency-domain projections, and autocorrelation structures—encode discriminative information that may not be easily recoverable from raw signals alone. While prior studies have explored representation ensembles or feature concatenation, systematic multi-representation learning within scalable deep architectures remains limited.

In parallel, multi-scale convolutional networks have proven effective for capturing temporal dependencies across varying receptive fields. Yet, current multi-scale models are typically optimized purely for predictive accuracy, with limited analysis of calibration quality and efficiency trade-offs. For large benchmark collections, such as the 142-dataset UCR archive\cite{dau2019ucr}, scalability and robustness become critical design considerations.

In this work, we propose a principled multi-representation multi-scale learning framework for TSC. Our contributions are threefold:

\begin{itemize}
    \item \textbf{Scalable Multi-Scale Architecture (MRMS-Net).} We introduce MRMS-Net, a hierarchical multi-scale convolutional network designed to integrate structured representation groups while maintaining stable calibration performance.
    
    \item \textbf{Lightweight Efficiency-Oriented Variant (LMRMS-Net).} We design LMRMS-Net as a computationally efficient alternative that preserves competitive predictive performance while significantly reducing training cost.LMRMS-Net incorporates a dynamic inference mechanism inspired by early-exit architectures \cite{teerapittayanon2016branchynet}. Unlike static models that apply uniform computation to all samples, LMRMS-Net employs a confidence-based gating strategy. By prioritizing shallow feature extraction for high-confidence samples and reserving the deeper fusion block for ambiguous cases, LMRMS-Net achieves a favorable trade-off between predictive latency and classification accuracy.
        
    \item \textbf{Multi-Representation Adaptation of LiteMV.} We repurpose LiteMV -originally developed for multivariate time series- to operate on structured multi-representation inputs of univariate signals, enabling cross-representation interaction through multivariate-style modeling.
\end{itemize}
            
We evaluate our models across 142 benchmark datasets under a unified experimental protocol with Monte Carlo resampling. Beyond reporting accuracy, we analyze macro-F1, Area Under the ROC Curve (AUC), negative log-likelihood (NLL), and runtime. Statistical validation using CD analysis confirms significant differences among the top-performing models.

Our results reveal three key findings. First, structured multi-representation learning consistently improves performance over raw inputs. Second, MRMS-Net achieves superior calibration performance, while LiteMV attains the highest overall accuracy. Third, LMRMS-Net establishes a strong efficiency–accuracy Pareto frontier, demonstrating that multi-scale modeling can be adapted to resource-constrained scenarios.

These findings establish scalable multi-representation multi-scale learning as a flexible and statistically validated paradigm for modern TSC.

\section{Related Work}

\subsection{Deep Learning for Time Series Classification}

Early TSC methods relied on distance-based approaches such as nearest neighbor classifiers \cite{lee2012nearest}  with elastic similarity measures. The introduction of deep learning shifted focus toward convolutional neural networks (CNNs), which demonstrated strong performance by automatically learning hierarchical temporal features from raw inputs. Architectures such as fully convolutional networks and residual networks became competitive baselines across large benchmark collections \cite{dhariyal2023back, shifaz2023elastic}.

More recently, attention-based \cite{wang2021new} transformer architectures \cite{le2024shapeformer} have further advanced TSC performance. However, many of these approaches prioritize predictive accuracy without explicitly addressing calibration quality or computational scalability across diverse dataset characteristics.

\subsection{Multi-Scale Modeling}
Early multi-scale approaches, such as the Multi-Scale Convolutional Neural Network (MCNN) \cite{cui2016multi} , introduced a transformation stage to extract multi-resolution features through down-sampling and smoothing. More recent multi-scale convolutional architectures \cite{ismail2020inceptiontime}  aim to capture temporal dependencies at different resolutions through parallel convolutional branches or hierarchical receptive fields. These designs have proven effective in modeling both short-term and long-term dynamics. Other CNN models \cite{cheng2021time} similarly stack dilated or multi-size filters to capture patterns of varying receptive fields. These architectures excel at accuracy, but rarely analyze their calibration or efficiency. Nonetheless, existing multi-scale networks generally operate on a single raw representation, implicitly assuming that scale diversity alone suffices to capture signal complexity. Beyond complex designs, Fully Convolutional Networks (FCNs) have demonstrated that superior performance can be achieved through relatively simple, parameter-efficient architectures that bypass pooling layers to preserve temporal resolution \cite{wang2017time}.

In contrast, our work combines scale diversity with representation diversity, enabling complementary information sources (e.g., time-domain derivatives, frequency magnitudes, autocorrelation) to be jointly modeled within a unified framework.

\subsection{Representation Learning and Multi-View Approaches}

Feature-based TSC approaches \cite{lubba2019catch22, middlehurst2022freshprince, christ2018time} have long leveraged handcrafted transformations such as wavelets \cite{li2016time}, Fourier coefficients \cite{schafer2016scalable}, and autocorrelation features to capture diverse temporal characteristics. Ensemble-based methods \cite{souza2025visemble, middlehurst2021hive} further combine heterogeneous representations at the classifier level to improve robustness and accuracy. 

Beyond individual feature sets, recent work has explored systematic integration of features extracted from multiple signal representations. For example, Crossfire \cite{alagoz2025crossfire} integrates features derived from derivative, autocorrelation, Fourier, cosine, wavelet, and Hilbert representations within a unified feature extraction framework. Evaluated on the 142 datasets of the UCR archive, this approach demonstrated that combining complementary representations can improve classification robustness while maintaining strong computational efficiency and scalability.

In parallel, convolutional kernel-based methods such as ROCKET and its variants \cite{dempster2020rocket, tan2022multirocket} have shown that generating large stochastic representation spaces using thousands of random convolutional kernels can effectively capture complex temporal patterns. Deep learning approaches have also been widely applied to TSC, often relying on ensembles of identical architectures trained independently to improve predictive performance \cite{abdullayev2025enhancing}. While ensembling improves accuracy, these models typically rely on random initialization to introduce diversity, which may lead to redundant feature representations.

More recently, representation learning paradigms have emerged to learn robust embeddings directly from time series data. Self-supervised methods such as TS2Vec \cite{yue2022ts2vec} and TF-C \cite{zhang2022self} aim to capture temporal and spectral dependencies through contrastive learning objectives. In parallel, multi-view learning frameworks have been explored in other domains to integrate complementary data sources and improve generalization.

Despite these advances, systematic integration of multiple signal representations within deep convolutional architectures remains relatively underexplored. In this work, we introduce multiple representation regimes and evaluate their impact across 142 datasets, providing large-scale empirical evidence that carefully designed representation combinations can yield consistent performance improvements.

\subsection{Multivariate Modeling and LiteMV}

LiteMV \cite{ismail2025look} was originally proposed for multivariate TSC, modeling interactions across channels. In this work, we reinterpret distinct signal representations as structured channels, allowing LiteMV to operate in a multi-representation setting. This adaptation enables cross-representation interaction without requiring inherently multivariate input signals, extending the applicability of multivariate architectures to representation-enhanced univariate problems.

\subsection{Calibration and Efficiency in TSC}

While accuracy remains the dominant evaluation metric in TSC, probabilistic calibration has gained increasing attention due to its importance in risk-sensitive applications. NLL provides a principled measure of predictive confidence quality. To this end, MRMS-Net is designed as a high-capacity, hierarchical architecture that leverages full representation diversity to achieve superior calibration and robustness across complex signal domains. 

Furthermore, large-scale empirical evaluations necessitate careful analysis of training and inference cost. The LITE model \cite{ismail2023lite} that accuracy-competitive CNN architectures for TSC can be achieved with significantly reduced parameter counts. Similarly, Omni-Scale architectures like OS-CNN \cite{tang2020omni} emphasize the importance of capturing universal patterns through diverse kernel sizes. Inspired by these efficiency-oriented design principles and neural scaling laws, LMRMS-Net incorporates lightweight convolutional strategies, such as reduced filter sizes and computationally efficient feature extraction, to maintain competitive performance while lowering computational cost.

Our study explicitly analyzes accuracy, macro-F1, AUC, NLL, and runtime, and visualizes trade-offs using Pareto analysis. To our knowledge, this is among the first works to jointly evaluate multi-scale, multi-representation architectures with statistical significance testing and efficiency–calibration tradeoff analysis across the full 142-dataset benchmark suite.

\section{Methodology}

\subsection{Multi-Representation Framework}

Rather than relying solely on raw time-domain signals, we construct structured representation sets designed to capture complementary temporal characteristics. For each univariate time series $x(t)$, we consider following representations: $TIME$, $DT1$, $DT2$, $HLB\_MAG$, $DWT\_A$, $FFT\_MAG$, $DCT$, and $ACF$.

Here, $DT1$ and $DT2$ denote first and second derivatives, $HLB\_MAG$ and $FFT\_MAG$ correspond to frequency magnitude projections, $DWT\_A$ represents wavelet approximation coefficients, $DCT$ denotes discrete cosine transform coefficients, and $ACF$ represents autocorrelation features.

Each representation is treated as an input channel, enabling structured multi-representation learning within convolutional architectures. This formulation allows controlled analysis of representation impact across datasets.

\subsection{Architectural Overview}

Figure~\ref{fig:architectures} provides a visual comparison between MRMS-Net and LMRMS-Net architectures. Both architectures process multi-representation inputs with shape $(R \times L)$, where $R$ is the number of representations and $L$ is the time series length.

\begin{figure*}[t]
\centering

\begin{minipage}{0.48\textwidth}
\centering

\resizebox{0.95\linewidth}{!}{
\begin{tikzpicture}[
node distance=1cm,
font=\scriptsize,
data/.style={cylinder, shape border rotate=90, draw=black!60, fill=gray!10,
minimum height=0.8cm, minimum width=2.8cm, aspect=0.25},
block/.style={rectangle, draw=green!70!black, fill=green!6, thick,
minimum width=3cm, minimum height=0.85cm, rounded corners, align=center},
conv/.style={rectangle, draw=green!70!black, fill=green!12, thick,
minimum width=2.4cm, minimum height=0.75cm, rounded corners, align=center},
arrow/.style={->, >=Stealth, semithick, draw=gray!80}
]

\node[data] (input) {Input $(R\times L)$};

\node[conv, below left=of input] (c3) {Conv1D $k=3$};
\node[conv, below=of input] (c5) {Conv1D $k=5$};
\node[conv, below right=of input] (c7) {Conv1D $k=7$};

\node[above=2pt of c3] {Short};
\node[above=2pt of c5] {Medium};
\node[above=2pt of c7] {Long};

\node[block, below=1.3cm of c5] (concat) {Concatenate};

\node[block, below=1.1cm of concat] (fusion)
{Feature Fusion\\BN → ReLU → Conv → Dropout};

\node[block, below=1.1cm of fusion]
(final) {Global Avg Pool + FC};

\draw[arrow] (input) -- (c3);
\draw[arrow] (input) -- (c5);
\draw[arrow] (input) -- (c7);

\draw[arrow] (c3) -- (concat);
\draw[arrow] (c5) -- (concat);
\draw[arrow] (c7) -- (concat);

\draw[arrow] (concat) -- (fusion);
\draw[arrow] (fusion) -- (final);

\end{tikzpicture}
}

\vspace{3pt}
\small (a) MRMS-Net architecture

\end{minipage}
\hfill
\begin{minipage}{0.48\textwidth}
\centering

\resizebox{0.95\linewidth}{!}{
\begin{tikzpicture}[
node distance=1cm,
font=\scriptsize,
data/.style={cylinder, shape border rotate=90, draw=black!60, fill=gray!10,
minimum height=0.8cm, minimum width=2.8cm, aspect=0.25},
block/.style={rectangle, draw=orange!70!black, fill=orange!6, thick,
minimum width=3cm, minimum height=0.85cm, rounded corners, align=center},
conv/.style={rectangle, draw=orange!70!black, fill=orange!12, thick,
minimum width=2.4cm, minimum height=0.75cm, rounded corners, align=center},
decision/.style={diamond, draw=orange!70!black, fill=orange!6, thick, aspect=2},
arrow/.style={->, >=Stealth, semithick, draw=gray!80}
]

\node[data] (input) {Input $(R\times L)$};

\node[conv, below left=of input] (c3) {Conv1D $k=3$};
\node[conv, below right=of input] (c5) {Conv1D $k=5$};

\node[block, below=1.4cm of $(c3)!0.5!(c5)$] (concat)
{Concatenate (32 ch)};

\node[decision, below=1.3cm of concat]
(dec) {$conf \ge \tau$};

\node[block, below left=of dec]
(exit) {Early Exit\\Pool + FC};

\node[block, below right=of dec]
(main) {Main Path\\Conv → Pool → FC};

\draw[arrow] (input) -- (c3);
\draw[arrow] (input) -- (c5);

\draw[arrow] (c3) -- (concat);
\draw[arrow] (c5) -- (concat);

\draw[arrow] (concat) -- (dec);

\draw[arrow] (dec) -- node[left, text=gray!70]{Yes} (exit);
\draw[arrow] (dec) -- node[right, text=gray!70]{No} (main);

\node[above=0.35cm of input,
draw=red!70!black,
fill=red!6,
rounded corners]
{Training: main path only};

\end{tikzpicture}
}

\vspace{3pt}
\small (b) LMRMS-Net architecture

\end{minipage}

\caption{
Architecture comparison of the proposed models.
MRMS-Net employs three multi-scale convolution branches ($k=3,5,7$) followed by a feature fusion block.
LMRMS-Net uses a lightweight two-branch design and incorporates a confidence-based early-exit mechanism to reduce inference cost.
}

\label{fig:architectures}

\end{figure*}
            
\subsection{MRMS-Net: Multi-Scale Representation Network}

MRMS-Net is designed to capture temporal dependencies at multiple receptive field scales while integrating structured representations.

Given an input tensor of shape $(R, L)$, where $R$ is the number of representations and $L$ is the series length, MRMS-Net applies parallel convolutional branches with different kernel sizes. These branches capture short-term and long-term temporal patterns simultaneously.

Branch outputs are concatenated and passed through hierarchical convolutional fusion blocks consisting of:

\begin{itemize}
    \item Batch normalization
    \item ReLU activation
    \item Stacked $1$D convolutions
    \item Dropout regularization
\end{itemize}

Global average pooling aggregates temporal information before classification. The architecture is optimized for stable training, controlled capacity growth, and robust calibration performance.

\subsection{LMRMS-Net: Lightweight Multi-Scale Network with Early Exit}

To address computational efficiency, we introduce LMRMS-Net (implemented as \texttt{FastMultiScaleCNN}), a lightweight multi-scale architecture with conditional early exit.

\subsubsection{Ultra-Light Multi-Scale Feature Extraction}

LMRMS-Net uses two shallow convolutional branches with kernel sizes $3$ and $5$:

\[
b_3 = \text{Conv1d}(R, 16, k=3), \quad
b_5 = \text{Conv1d}(R, 16, k=5)
\]

The branch outputs are concatenated to form a compact 32-channel representation.

\subsubsection{Early Exit Classifier}

An early classifier operates directly on pooled branch features:

\begin{itemize}
    \item Adaptive average pooling
    \item Fully connected layer (32 $\rightarrow$ 64)
    \item ReLU activation
    \item Output layer (64 $\rightarrow$ $C$)
\end{itemize}

During inference, prediction confidence is computed via softmax probabilities. If the mean maximum class probability exceeds a threshold $\tau = 0.8$, the early prediction is returned.

\subsubsection{Main Pathway (Fallback)}

If confidence is below threshold, features are processed through a deeper fusion block:

\begin{itemize}
    \item BatchNorm + ReLU
    \item Conv1d(32 $\rightarrow$ 64)
    \item Conv1d(64 $\rightarrow$ 128)
    \item Dropout (0.3)
\end{itemize}

After global average pooling, a final linear classifier produces predictions.

During training, only the main pathway is used to ensure stable gradient flow. Early exit is activated only during inference.

This design enables LMRMS-Net to reduce inference cost on ``easy'' samples while maintaining competitive accuracy.

\subsection{LiteMV Multi-Representation Adaptation}

LiteMV was originally designed for multivariate TSC. We reinterpret representation channels as structured pseudo-variables, enabling cross-representation interaction modeling.

Formally, for a representation set of size $R$, the input tensor is treated as multivariate with $R$ channels. LiteMV thus models:

\[
\mathcal{F}: \mathbb{R}^{R \times L} \rightarrow \mathbb{R}^{C}
\]

This adaptation allows structured interaction between time-domain and frequency-domain signals without requiring inherently multivariate datasets.

\section{Experimental Protocol}

\subsection{Datasets and Evaluation}

We evaluate all models on 142 benchmark TSC datasets. For each dataset, we employ Monte Carlo resampling with $R$ repeated train/test splits.

Predefined resampling indices are used when available to ensure strict comparability with prior state-of-the-art (SOTA) studies.

Performance is reported as:
\begin{itemize}
    \item Mean across resamples (per dataset),
    \item Then macro-averaged across datasets.
\end{itemize}

This avoids dataset-size bias and follows established large-scale evaluation protocols.

\subsection{Training Configuration}

All models are trained using the Adam optimizer with cross-entropy loss for a maximum of 1500 epochs.

Early stopping is applied based on \textbf{training loss} with a fixed patience parameter. The best model state is restored before evaluation.

Batch size is automatically selected as a function of dataset workload ($N \times L$), with dynamic adjustment to prevent GPU out-of-memory failures.

\subsection{Evaluation Metrics}

For each resample, we compute Accuracy, Macro F1-score, AUC (computed for both binary and multi-class cases), NLL, and training and test time.

Final rankings and statistical comparisons are conducted using the Friedman test with Nemenyi post-hoc analysis.

\section{Results}
\label{sec:results}

We evaluate four primary architectures across 142 UCR/UEA datasets using 30 Monte-Carlo resamples per dataset. Performance is measured using accuracy, macro-F1, AUC, and NLL. Training and test times are also recorded to assess computational efficiency.

The four architectures compared in detail are:

\begin{itemize}
    \item Lite (baseline)
    \item LiteMV (multi-view adaptation)
    \item LMRMS-Net (Lightweight Scale Network)
    \item MRMS-Net (Multi-Scale Network)
\end{itemize}

All statistical comparisons are performed using the Friedman test followed by Nemenyi post-hoc analysis across 142 datasets.

\subsection{Overall Performance Comparison}
\label{subsec:overall_performance}

Table~\ref{tab:main_results} reports mean performance across all datasets.

\begin{table*}[t]
\centering
\caption{Mean performance across 142 datasets (best values in bold).}
\label{tab:main_results}
\begin{tabular}{lcccccc}
\toprule
Architecture & Accuracy & F1 & AUC & NLL $\downarrow$ & Train Time (s) & Test Time (s) \\
\midrule
Lite & 0.828 & 0.802 & 0.936 & 0.675 & 21.26 & 0.059 \\
LiteMV & \textbf{0.836} & \textbf{0.812} & 0.938 & 0.647 & 18.72 & 0.088 \\
LMRMS-Net & 0.827 & 0.801 & 0.939 & 0.677 & \textbf{11.70} & \textbf{0.027} \\
MRMS-Net & 0.828 & 0.799 & \textbf{0.938} & \textbf{0.615} & 25.35 & 0.088 \\
\bottomrule
\end{tabular}
\end{table*}

LiteMV achieves the highest mean accuracy and macro-F1, while MRMS-Net achieves the best calibration (lowest NLL). LMRMS-Net provides competitive accuracy with significantly reduced computational cost.

\subsection{Statistical Significance Across 142 Datasets}
\label{subsec:statistical}

\begin{figure}[t]
    \centering
    \includegraphics[width=0.75\linewidth]{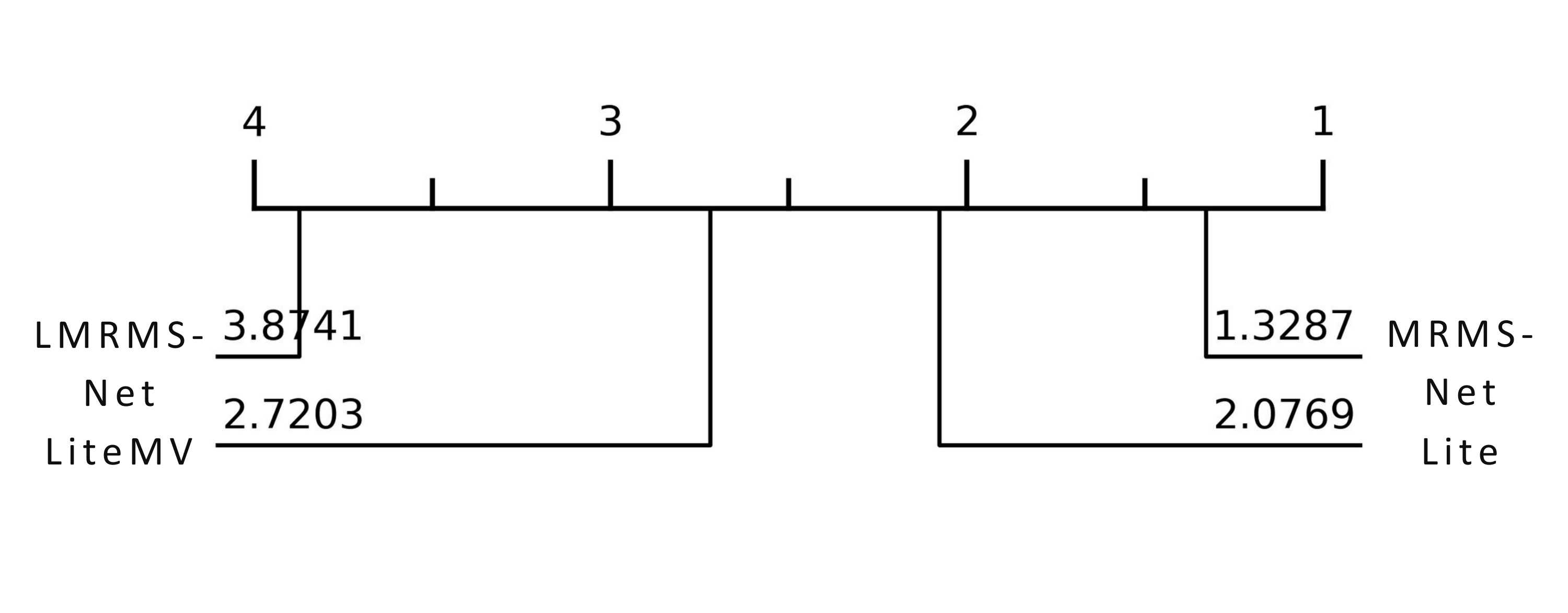}
    \caption{CD diagram based on accuracy rankings across 142 datasets. Lower ranks indicate better performance. Methods connected by a horizontal bar are not significantly different according to the Nemenyi test.}
    \label{fig:cd}
\end{figure}

Figure~\ref{fig:cd} presents the CD diagram based on accuracy rankings across 142 datasets.

The Friedman test indicates statistically significant differences among methods ($p < 0.05$). The Nemenyi post-hoc test shows: (i) LiteMV ranks first overall, (ii) Lite and MRMS-Net are statistically indistinguishable from LiteMV, and (iii) LMRMS-Net remains competitive but slightly lower in average rank.

Importantly, no architecture dominates all others across every dataset, confirming that improvements are dataset-dependent.

\begin{figure}[t]
    \centering
    \includegraphics[width=\linewidth]{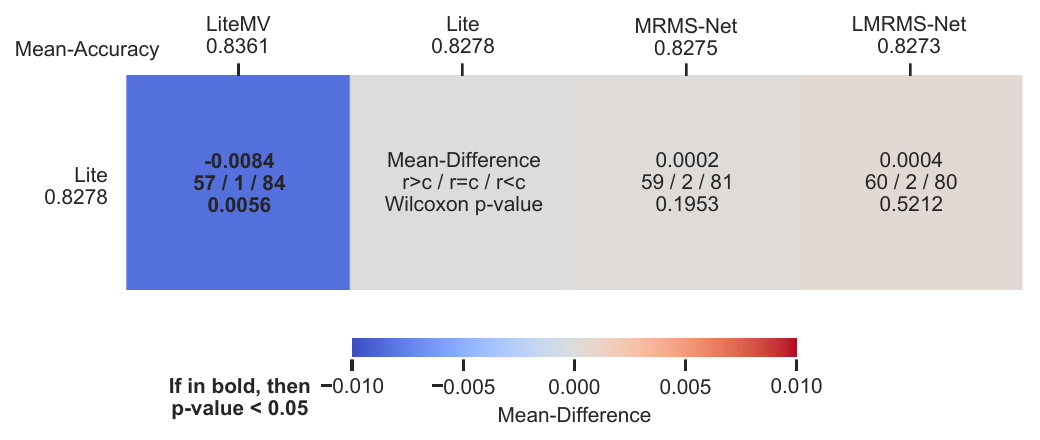}
    \caption{Multi comparision matrix.}
    \label{fig:mcm}
\end{figure}
\subsection{Efficiency–Performance Tradeoff}
\label{subsec:pareto}

\begin{figure}[t]
    \centering
    \includegraphics[width=0.85\linewidth]{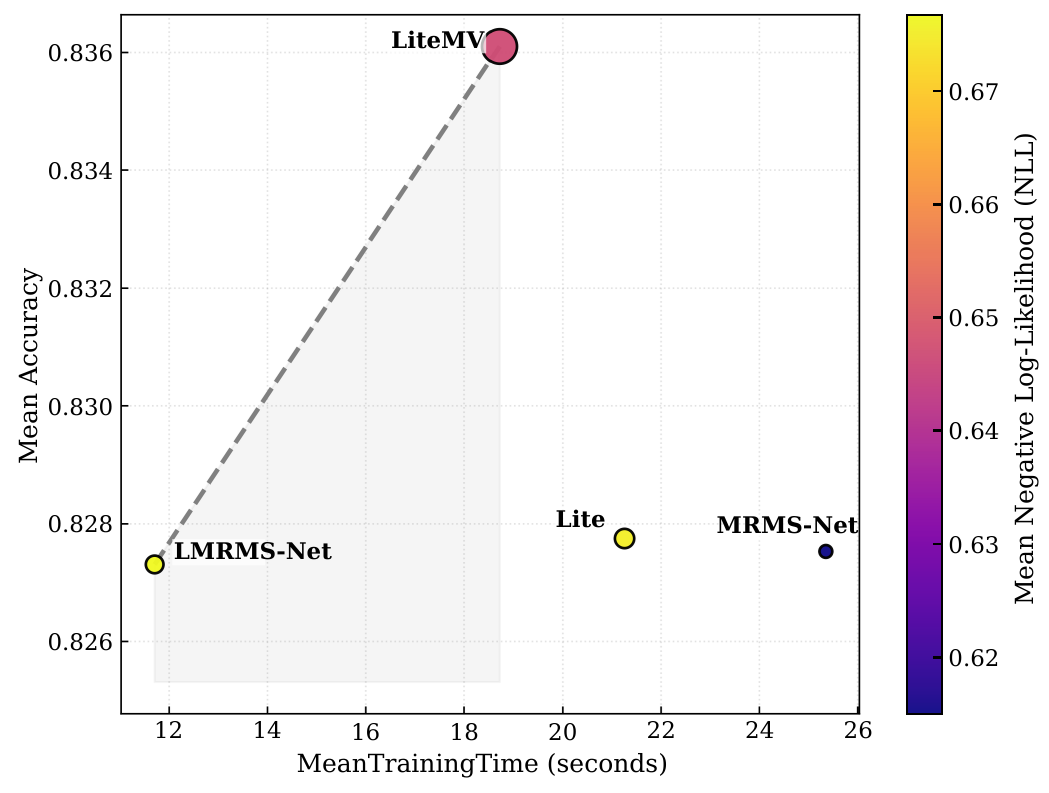}
    \caption{Pareto tradeoff between mean training time and classification accuracy. Marker size represents mean AUC, while color encodes mean NLL. The dashed curve denotes the Pareto frontier, identifying models that achieve optimal tradeoffs between predictive performance and computational cost.}
    \label{fig:pareto}
\end{figure}

Figure~\ref{fig:pareto} shows the Pareto tradeoff between mean training time and accuracy. Marker size represents AUC, and color encodes NLL.
LMRMS-Net lies near the Pareto frontier, achieving near-SOTA accuracy with substantially reduced training time. LiteMV provides the strongest accuracy while maintaining moderate training cost. MRMS-Net achieves superior calibration but at increased computational expense.

This demonstrates that scalable multi-scale modeling can be adapted to different operating regimes:
\begin{itemize}
    \item \textbf{Accuracy-oriented regime:} LiteMV
    \item \textbf{Efficiency-oriented regime:} LMRMS-Net
    \item \textbf{Calibration-oriented regime:} MRMS-Net
\end{itemize}

\subsection{Calibration Analysis}
\label{subsec:calibration}

\begin{figure}[t]
    \centering
    \includegraphics[width=0.85\linewidth]{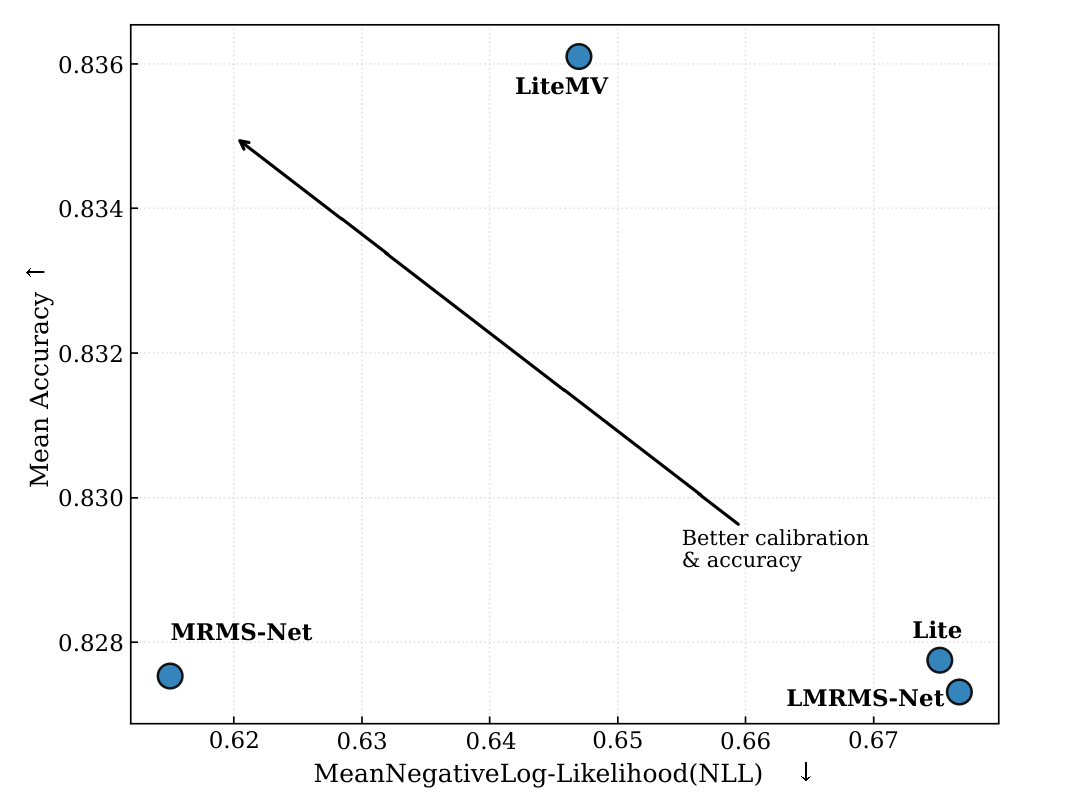}
    \caption{Accuracy versus mean NLL across evaluated architectures. }
    \label{fig:calib}
\end{figure}

Figure~\ref{fig:calib} plots accuracy versus NLL. While several models achieve similar accuracy, MRMS-Net consistently achieves lower NLL values, indicating better probabilistic calibration.
This suggests that multi-scale feature aggregation contributes not only to classification accuracy but also to improved uncertainty estimation.

\subsection{Impact of Representations}
\label{subsec:representation}

\subsubsection{Architecture–Representation Interaction}

The benefit of representation expansion varies across architectures:

\begin{itemize}
    \item LiteMV benefits most strongly, likely due to cross-channel interaction.
    \item MRMS-Net shows stable improvements, indicating inherent robustness to representation diversity.
    \item LMRMS-Net achieves optimal efficiency under the Minimal setting.
\end{itemize}

These findings demonstrate that representation diversity interacts with architectural design in non-trivial ways.

\subsubsection{Efficiency Considerations}

Although the Default representation often yields the highest accuracy, it increases training time. The Minimal set captures most gains at significantly lower computational cost, offering a strong tradeoff point.

\subsection{Summary of Findings}

Across 142 datasets, the results support three main conclusions:

\begin{enumerate}
    \item Multi-view representation expansion significantly improves performance over raw time-domain inputs.
    \item Scalable multi-scale convolution provides strong calibration benefits.
    \item Lightweight variants can achieve near state-of-the-art performance with substantially reduced computational cost.
\end{enumerate}

Overall, combining structured representation diversity with scalable multi-scale architectures forms a robust and efficient framework for TSC.

\section{Discussion}
\label{sec:discussion}

This study provides large-scale empirical evidence across 142 datasets that performance in TSC is governed by three interacting factors: architectural capacity, representation diversity, and computational scalability.

\subsection{Architecture vs. Representation}

A key finding is that representation diversity consistently improves performance across all architectures. Moving from raw time-domain inputs to the Minimal representation set yields substantial gains in both accuracy and macro-F1. Expanding further to the Default set provides smaller but consistent improvements, suggesting diminishing returns beyond a compact, informative transformation core.

Interestingly, the magnitude of improvement depends on architectural design. LiteMV benefits the most from representation expansion, indicating that cross-view interactions effectively exploit complementary feature domains. In contrast, MRMS-Net exhibits more stable performance across representation regimes, suggesting inherent robustness due to multi-scale aggregation. LMRMS-Net achieves its best efficiency–accuracy balance under the Minimal setting, indicating that lightweight models benefit most from carefully curated representation subsets.

These results highlight that representation engineering and architectural design should not be treated independently; their interaction is central to scalable TSC performance.

\subsection{Accuracy vs. Calibration}

While LiteMV achieves the highest mean accuracy, MRMS-Net consistently attains the lowest NLL, indicating superior probabilistic calibration. This suggests that multi-scale hierarchical aggregation improves uncertainty estimation beyond pure classification accuracy.

This distinction is important for applications requiring reliable confidence estimates, such as medical diagnosis or anomaly detection. The results imply that architectural depth and multi-scale structure contribute differently to discrimination and calibration.

\subsection{Efficiency Considerations}

From an efficiency standpoint, LMRMS-Net demonstrates that competitive accuracy can be achieved with substantially reduced training and inference cost. The Pareto analysis shows that LMRMS-Net lies near the efficiency frontier, making it attractive for large-scale or resource-constrained deployments.

Importantly, the representation expansion strategy does not break scalability. The Minimal representation captures most performance gains while preserving computational efficiency, making it a strong default configuration for practical applications.

\subsection{Statistical Robustness}

The Friedman and Nemenyi analyses confirm statistically significant differences among architectures. However, no single model dominates across all datasets. This reinforces the importance of reporting average ranks and performing multi-dataset statistical testing rather than relying solely on mean accuracy.

Overall, the results demonstrate that scalable multi-scale convolution combined with structured multi-representation inputs forms a robust and adaptable TSC framework.
\section{Conclusion}
\label{sec:conclusion}

We introduced a scalable multi-scale convolutional framework for TSC and systematically evaluated its behavior across 142 benchmark datasets.

Our contributions can be summarized as follows:

\begin{enumerate}
    \item We demonstrated that structured representation expansion (Raw → Minimal → Default) consistently improves classification performance.
    \item We showed that adapting LiteMV to multi-representation univariate inputs provides strong accuracy gains.
    \item We proposed MRMS-Net and LMRMS-Net, scalable multi-scale architectures that balance accuracy, calibration, and computational efficiency.
    \item We provided statistically rigorous comparisons using CD analysis and Pareto tradeoff evaluation.
\end{enumerate}

The results reveal that:
\begin{itemize}
    \item LiteMV achieves the highest mean accuracy across 142 datasets.
    \item MRMS-Net provides superior calibration performance.
    \item LMRMS-Net achieves competitive accuracy with significantly reduced training cost.
\end{itemize}

These findings suggest that combining representation diversity with scalable multi-scale modeling offers a flexible design space that can be tuned for accuracy-oriented, calibration-oriented, or efficiency-oriented regimes.

Future work will investigate adaptive representation selection, dynamic multi-scale attention mechanisms, and extension to large-scale multivariate benchmarks.

Overall, this study establishes that scalable multi-representation multi-scale learning is a principled and practical direction for modern TSC.

\bibliographystyle{IEEEtran}
\bibliography{refs}

@inproceedings{dhariyal2023back,
  title={Back to basics: A sanity check on modern time series classification algorithms},
  author={Dhariyal, Bhaskar and Le Nguyen, Thach and Ifrim, Georgiana},
  booktitle={International Workshop on Advanced Analytics and Learning on Temporal Data},
  pages={205--229},
  year={2023},
  organization={Springer}
}

@article{shifaz2023elastic,
  title={Elastic similarity and distance measures for multivariate time series},
  author={Shifaz, Ahmed and Pelletier, Charlotte and Petitjean, Fran{\c{c}}ois and Webb, Geoffrey I},
  journal={Knowledge and Information Systems},
  volume={65},
  number={6},
  pages={2665--2698},
  year={2023},
  publisher={Springer}
}

@article{lee2012nearest,
  title={Nearest-neighbor-based approach to time-series classification},
  author={Lee, Yen-Hsien and Wei, Chih-Ping and Cheng, Tsang-Hsiang and Yang, Ching-Ting},
  journal={Decision Support Systems},
  volume={53},
  number={1},
  pages={207--217},
  year={2012},
  publisher={Elsevier}
}

@article{wang2021new,
  title={A new attention-based CNN approach for crop mapping using time series Sentinel-2 images},
  author={Wang, Yumiao and Zhang, Zhou and Feng, Luwei and Ma, Yuchi and Du, Qingyun},
  journal={Computers and electronics in agriculture},
  volume={184},
  pages={106090},
  year={2021},
  publisher={Elsevier}
}

@inproceedings{le2024shapeformer,
  title={Shapeformer: Shapelet transformer for multivariate time series classification},
  author={Le, Xuan-May and Luo, Ling and Aickelin, Uwe and Tran, Minh-Tuan},
  booktitle={Proceedings of the 30th ACM SIGKDD Conference on Knowledge Discovery and Data Mining},
  pages={1484--1494},
  year={2024}
}

@article{dau2019ucr,
  title={The UCR time series archive},
  author={Dau, Hoang Anh and Bagnall, Anthony and Kamgar, Kaveh and Yeh, Chin-Chia Michael and Zhu, Yan and Gharghabi, Shaghayegh and Ratanamahatana, Chotirat Ann and Keogh, Eamonn},
  journal={IEEE/CAA Journal of Automatica Sinica},
  volume={6},
  number={6},
  pages={1293--1305},
  year={2019},
  publisher={IEEE}
}

@article{ismail2025look,
  title={Look into the lite in deep learning for time series classification},
  author={Ismail-Fawaz, Ali and Devanne, Maxime and Berretti, Stefano and Weber, Jonathan and Forestier, Germain},
  journal={International Journal of Data Science and Analytics},
  volume={20},
  number={4},
  pages={4029--4049},
  year={2025},
  publisher={Springer}
}

@article{lubba2019catch22,
  title={catch22: CAnonical Time-series CHaracteristics: Selected through highly comparative time-series analysis},
  author={Lubba, Carl H and Sethi, Sarab S and Knaute, Philip and Schultz, Simon R and Fulcher, Ben D and Jones, Nick S},
  journal={Data mining and knowledge discovery},
  volume={33},
  number={6},
  pages={1821--1852},
  year={2019},
  publisher={Springer}
}

@inproceedings{middlehurst2022freshprince,
  title={The freshprince: A simple transformation based pipeline time series classifier},
  author={Middlehurst, Matthew and Bagnall, Anthony},
  booktitle={International Conference on Pattern Recognition and Artificial Intelligence},
  pages={150--161},
  year={2022},
  organization={Springer}
}

@article{christ2018time,
  title={Time series feature extraction on basis of scalable hypothesis tests (tsfresh--a python package)},
  author={Christ, Maximilian and Braun, Nils and Neuffer, Julius and Kempa-Liehr, Andreas W},
  journal={Neurocomputing},
  volume={307},
  pages={72--77},
  year={2018},
  publisher={Elsevier}
}

@article{schafer2016scalable,
  title={Scalable time series classification},
  author={Sch{\"a}fer, Patrick},
  journal={Data Mining and Knowledge Discovery},
  volume={30},
  number={5},
  pages={1273--1298},
  year={2016},
  publisher={Springer}
}

@article{li2016time,
  title={Time series classification with discrete wavelet transformed data},
  author={Li, Daoyuan and Bissyande, Tegawende F and Klein, Jacques and Traon, Yves Le},
  journal={International Journal of Software Engineering and Knowledge Engineering},
  volume={26},
  number={09n10},
  pages={1361--1377},
  year={2016},
  publisher={World Scientific}
}

@article{souza2025visemble,
  title={Visemble: A fast ensemble approach for time series classification with multiple visual representations},
  author={Souza, Vinicius MA and Veiga, Patrickerson S and Ribeiro, Andr{\'e} GR},
  journal={Knowledge-Based Systems},
  volume={309},
  pages={112864},
  year={2025},
  publisher={Elsevier}
}

@article{middlehurst2021hive,
  title={HIVE-COTE 2.0: a new meta ensemble for time series classification},
  author={Middlehurst, Matthew and Large, James and Flynn, Michael and Lines, Jason and Bostrom, Aaron and Bagnall, Anthony},
  journal={Machine Learning},
  volume={110},
  number={11},
  pages={3211--3243},
  year={2021},
  publisher={Springer}
}

@inproceedings{abdullayev2025enhancing,
  title={Enhancing Time Series Classification with Diversity-Driven Neural Network Ensembles},
  author={Abdullayev, Javidan and Devanne, Maxime and Meyer, Cyril and Ismail-Fawaz, Ali and Weber, Jonathan and Forestier, Germain},
  booktitle={2025 International Joint Conference on Neural Networks (IJCNN)},
  pages={1--10},
  year={2025},
  organization={IEEE}
}

@article{ismail2020inceptiontime,
  title={Inceptiontime: Finding alexnet for time series classification},
  author={Ismail Fawaz, Hassan and Lucas, Benjamin and Forestier, Germain and Pelletier, Charlotte and Schmidt, Daniel F and Weber, Jonathan and Webb, Geoffrey I and Idoumghar, Lhassane and Muller, Pierre-Alain and Petitjean, Fran{\c{c}}ois},
  journal={Data mining and knowledge discovery},
  volume={34},
  number={6},
  pages={1936--1962},
  year={2020},
  publisher={Springer}
}

@article{cheng2021time,
  title={Time series classification using diversified ensemble deep random vector functional link and resnet features},
  author={Cheng, Wen Xin and Suganthan, Ponnuthurai N and Katuwal, Rakesh},
  journal={Applied Soft Computing},
  volume={112},
  pages={107826},
  year={2021},
  publisher={Elsevier}
}

@inproceedings{ismail2023lite,
  title={Lite: Light inception with boosting techniques for time series classification},
  author={Ismail Fawaz, Ali and Devanne, Maxime and Berretti, Stefano and Weber, Jonathan and Forestier, Germain},
  booktitle={2023 IEEE 10th International Conference on Data Science and Advanced Analytics (DSAA)},
  pages={1--10},
  year={2023},
  organization={IEEE}
}

@article{cui2016multi,
  title={Multi-scale convolutional neural networks for time series classification},
  author={Cui, Zhicheng and Chen, Wenlin and Chen, Yixin},
  journal={arXiv preprint arXiv:1603.06995},
  year={2016}
}

@inproceedings{wang2017time,
  title={Time series classification from scratch with deep neural networks: A strong baseline},
  author={Wang, Zhiguang and Yan, Weizhong and Oates, Tim},
  booktitle={2017 International joint conference on neural networks (IJCNN)},
  pages={1578--1585},
  year={2017},
  organization={IEEE}
}

@article{dempster2020rocket,
  title={ROCKET: exceptionally fast and accurate time series classification using random convolutional kernels},
  author={Dempster, Angus and Petitjean, Fran{\c{c}}ois and Webb, Geoffrey I},
  journal={Data Mining and Knowledge Discovery},
  volume={34},
  number={5},
  pages={1454--1495},
  year={2020},
  publisher={Springer}
}

@article{tan2022multirocket,
  title={MultiRocket: multiple pooling operators and transformations for fast and effective time series classification: CW Tan},
  author={Tan, Chang Wei and Dempster, Angus and Bergmeir, Christoph and Webb, Geoffrey I},
  journal={Data Mining and Knowledge Discovery},
  volume={36},
  number={5},
  pages={1623--1646},
  year={2022},
  publisher={Springer}
}

@inproceedings{yue2022ts2vec,
  title={Ts2vec: Towards universal representation of time series},
  author={Yue, Zhihan and Wang, Yujing and Duan, Juanyong and Yang, Tianmeng and Huang, Congrui and Tong, Yunhai and Xu, Bixiong},
  booktitle={Proceedings of the AAAI conference on artificial intelligence},
  volume={36},
  number={8},
  pages={8980--8987},
  year={2022}
}

@article{zhang2022self,
  title={Self-supervised contrastive pre-training for time series via time-frequency consistency},
  author={Zhang, Xiang and Zhao, Ziyuan and Tsiligkaridis, Theodoros and Zitnik, Marinka},
  journal={Advances in neural information processing systems},
  volume={35},
  pages={3988--4003},
  year={2022}
}

@inproceedings{teerapittayanon2016branchynet,
  title={Branchynet: Fast inference via early exiting from deep neural networks},
  author={Teerapittayanon, Surat and McDanel, Bradley and Kung, Hsiang-Tsung},
  booktitle={2016 23rd international conference on pattern recognition (ICPR)},
  pages={2464--2469},
  year={2016},
  organization={IEEE}

}

@article{tang2020omni,
  title={Omni-scale cnns: a simple and effective kernel size configuration for time series classification},
  author={Tang, Wensi and Long, Guodong and Liu, Lu and Zhou, Tianyi and Blumenstein, Michael and Jiang, Jing},
  journal={arXiv preprint arXiv:2002.10061},
  year={2020}
}

@article{alagoz2025crossfire,
  title={Crossfire: cross-domain feature integration for robust time series classification},
  author={Alag{\"o}z, Celal},
  journal={PeerJ Computer Science},
  volume={11},
  pages={e3328},
  year={2025},
  publisher={PeerJ Inc.}
}

\end{document}